\documentclass{article}

\usepackage[final]{nips_2017}

\usepackage[utf8]{inputenc} % allow utf-8 input
\usepackage[T1]{fontenc}    % use 8-bit T1 fonts
\usepackage{hyperref}       % hyperlinks
\usepackage{url}            % simple URL typesetting
\usepackage{booktabs}       % professional-quality tables
\usepackage{amsfonts}       % blackboard math symbols
\usepackage{nicefrac}       % compact symbols for 1/2, etc.
\usepackage{microtype}      % microtypography
\usepackage{amssymb}
\usepackage{graphicx}
\usepackage{amsmath}
\usepackage{latexsym}
\usepackage{mathtools}
\usepackage{bm}
\usepackage{listings}
\usepackage{subfig}
\usepackage{array}
\usepackage{caption}
\usepackage[pdftex,dvipsnames]{xcolor}
\usepackage[colorinlistoftodos,prependcaption,textsize=tiny]{todonotes}

\usepackage{comment}

\newcommand{\vect}[1]{\mathbf{#1}}

\newcommand{\matr}[1]{\mathbf{#1}}

\newcommand{\va}[0]{\vect{a}}
\newcommand{\vb}[0]{\vect{b}}
\newcommand{\vc}[0]{\vect{c}}

\newcommand{\vg}[0]{\vect{g}}
\newcommand{\vh}[0]{\vect{h}}

\newcommand{\vr}[0]{\vect{r}}

\newcommand{\vv}[0]{\vect{v}}

\newcommand{\vx}[0]{\vect{x}}
\newcommand{\vy}[0]{\vect{y}}
\newcommand{\vz}[0]{\vect{z}}

\newcommand{\mM}[0]{\matr{M}}

\newcommand{\mP}[0]{\matr{P}}

\newcommand{\mR}[0]{\matr{R}}

\newcommand{\mW}[0]{\matr{W}}

\newcommand{\mZ}[0]{\matr{Z}}

\allowdisplaybreaks  % remove this if equations look awkwardly split over pages

\graphicspath{{./figures/}}

\newcommand*{\affaddr}[1]{#1} % No op here. Customize it for different styles.
\newcommand*{\affmark}[1][*]{\textsuperscript{#1}}
\newcommand*{\email}[1]{\texttt{#1}}
\newcommand\tab[1][1cm]{\hspace*{#1}}

\title{Grounded Recurrent Neural Networks}

\author{%
Ankit Vani\affmark[$\ast$] \tab Yacine Jernite\affmark[$\dagger$] \tab David Sontag\affmark[$\ddagger$]\\
\vspace{-0.12in}\\
\affaddr{\affmark[$\ast \dagger$]CIMS, New York University, New York, NY 10012}\\
\affaddr{\affmark[$\ddagger$]CSAIL \& IMES, Massachusetts Institute of Technology, Cambridge, MA 02139}\\
\affmark[$\ast$]\email{ankit.vani@nyu.edu},\, \affmark[$\dagger$]\email{jernite@cs.nyu.edu},\, \affmark[$\ddagger$]\email{dsontag@csail.mit.edu}\\
}

\begin{document}
% \nipsfinalcopy is no longer used

\maketitle

%Main points:
%
%- Recurrent state explicitly models beliefs for each label
%    - Why?
%        - Easier optimization
%        - By dedicating certain dimensions to certain labels, allowing the model to condition on explicit disentangled features that the model is more confident  on
%        - Allows the model to more directly model correlations between evolving states as text is read
%    - Grounded RNN
%        - How is the invariant maintained at every timestep?
%            - Curriculum learning helps
%            - Empirically (show localization visualization)

\begin{abstract}
  In this work, we present the Grounded Recurrent Neural Network (GRNN), a recurrent neural network architecture for multi-label prediction which explicitly ties labels to specific dimensions of the recurrent hidden state (we call this process ``grounding''). The approach is particularly well-suited for extracting large numbers of concepts from text. We apply the new model to address an important problem in healthcare of understanding what medical concepts are discussed in clinical text. Using a publicly available dataset derived from Intensive Care Units, we learn to label a patient's diagnoses and procedures from their discharge summary. Our evaluation shows a clear advantage to using our proposed architecture over a variety of strong baselines.
\end{abstract}

\section{Introduction}

The ability of recurrent neural networks to model sequential data and capture long-term dependencies makes them powerful tools for natural language processing. These models maintain a state at each time step, representing the relevant history and task-specific beliefs. Based on the current value of this recurrent state and a new input, the state is updated at each time step. Recurrent models have become a popular choice for a variety of natural language processing tasks such as language modeling \citep{DBLP:conf/interspeech/MikolovKBCK10}, text classification \citep{DBLP:series/sci/2012-385}, or machine translation \citep{DBLP:conf/ssst/ChoMBB14}. The success of this paradigm has been driven in great part by a number of structural innovations since the original version of \cite{DBLP:journals/cogsci/Elman90}. Recurrent cells such as the Long Short Term Memory (LSTM) \citep{lstm} or Gated Recurrent Units (GRU) \citep{gru}, for example, alleviate the problem of vanishing gradients \citep{vanishing-grad}. Attention mechanisms \citep{DBLP:journals/corr/BahdanauCB14} and Memory Networks \citep{DBLP:conf/nips/SukhbaatarSWF15} have also significantly increased the expressiveness of recurrent architectures, revealing their potential to tackle more complex tasks such as question answering \citep{DBLP:conf/emnlp/RajpurkarZLL16}. One notable property of these models, however, is that they often require significant amounts of training data to perform at their best \citep{DBLP:journals/corr/BajgarKK16}, which can limit their application domain.

In this work, we focus on developing recurrent models for the task of extracting medical concepts from Intensive Care Unit discharge summaries. This is a multi-class, multi-label text classification task with a target vocabulary of several thousand concepts. Given the difficulty to obtain very large medical datasets, there is a need to come up with new, more data-efficient architectures. To this end, we introduce the Grounded Recurrent Neural Network (GRNN). At a high level, we \emph{ground} the model's hidden state by introducing dimensions whose sole purpose is to track the model's belief in the presence of specific labels for the current example. We find that this addition aids optimization, and outperforms standard recurrent models for text classification. Although each new label adds to the hidden state size, we impose a semi diagonal constraint on the recurrent transition matrices, so that the size of our model grows linearly with the number of labels. We show that this not only lets the model scale with the number of labels, but also helps with optimization. Furthermore, our approach leads to increased interpretability, which is especially appreciated in medical applications. Indeed, even though we do not provide our model with the location of phrases of interest for a label at training time, we can track changes in the grounded dimensions tied to a specific concept as a document is read, indicating evidence in text for or against its presence when the model's belief changes drastically.

We evaluate our model on the publicly available MIMIC datasets \citep{mimic2,mimic3} to predict ICD9 (International Classification of Diseases \citep{icd9}) codes given a patient's discharge summary text \citep{baseline,lita2008large}. These codes are usually determined by humans perusing health records and selecting relevant codes from very long lists. Due to the high number of ICD9 codes, there is significant human error, arising from the cognitive load of such a task \citep{birman2005accuracy,hsia1988accuracy} and differences in human judgment \citep{Pestian:2007:STI:1572392.1572411}. The effort needed, the errors in the coding process and the inconsistency of labeling can be mitigated through automatically detecting concepts in text or offering suggestions as smarter auto-complete, which motivates our contribution. We also show our model's performance on a tag prediction dataset built from StackOverflow data.

Section \ref{sec:related} presents relevant previous work, Section \ref{sec:grnn} describes the model, and we present experimental results in Section \ref{sec:experiments}. Section \ref{sec:conclusion} concludes and outlines possible future research directions.

\section{Related Work}
\label{sec:related}

\paragraph{Entity Tracking} The idea to keep information relating to specific concepts in dedicated cells was inspired in part by the work of \cite{henaff2016tracking} on recurrent entity networks. They propose to define one full RNN per entity of interest which is tasked with keeping track of all of the relevant information as a story is read, and their model is able to answer questions about the state of the world by consulting all of their final hidden states. Such an approach is unfortunately impractical for our case, which presents thousands of concepts of interest. Instead, we use one dimension in the recurrent hidden state for each of them to track one specific piece of information: the likelihood that they are present in the current example.

\paragraph{Sparse Recurrent Units} \cite{subakan2017diagonal} use diagonal recurrent transition matrices for music modeling and show that their model outperforms the the one with full transition matrices. The upper-left block of the transition matrix in our model is diagonal. \cite{narang2017exploring} explored sparsity in recurrent models, and showed that sparse weight matrices result in much smaller models without significant loss in accuracy. The weight sparsity introduced in our model through a semi diagonal weight matrix additionally limits over-fitting, as discussed in Section \ref{sec:grnn}.

\paragraph{Interpretable RNNs} \cite{lei2016rationalizing} propose a method for encouraging interpretability in the learning objective by selecting predictive phrases as the first step. Additionally, \cite{lample2016neural} discuss recurrent models followed by a pairwise conditional random field (CRF) for named entity recognition. Their LSTM+CRF model is used to predict word tags that determine the span a word belongs to and the entity the span represents. They also present a transition-based chunking model which uses a stacked LSTM where words are pushed onto a stack and popped with an entity label. 
%By grounding dimensions to labels, we find that such visualizations are more reliable than using the final state label projection on intermediate states in standard recurrent networks.

\paragraph{Medical Concept Extraction} \cite{joshi2016identifiable} use non-negative matrix factorization for simultaneous phenotyping of co-occurring medical conditions from clinical text. They obtain identifiable sparse latent factors by grounding them to have a one-to-one mapping with a fixed set of chronic conditions. Our model grounds its recurrent hidden state dimensions to have a one-to-one mapping with the labels for a task. Automatic ICD9 coding has been explored in clinical machine learning literature, such as \cite{baseline}. Their work presents a tree structured Support Vector Machine (SVM) which takes advantage of the hierarchical nature of ICD9 codes to outperform a flat baseline.

\section{Grounded Recurrent Neural Networks}
\label{sec:grnn}

\begin{figure}[t!]
    \centering
    \includegraphics[width=\textwidth]{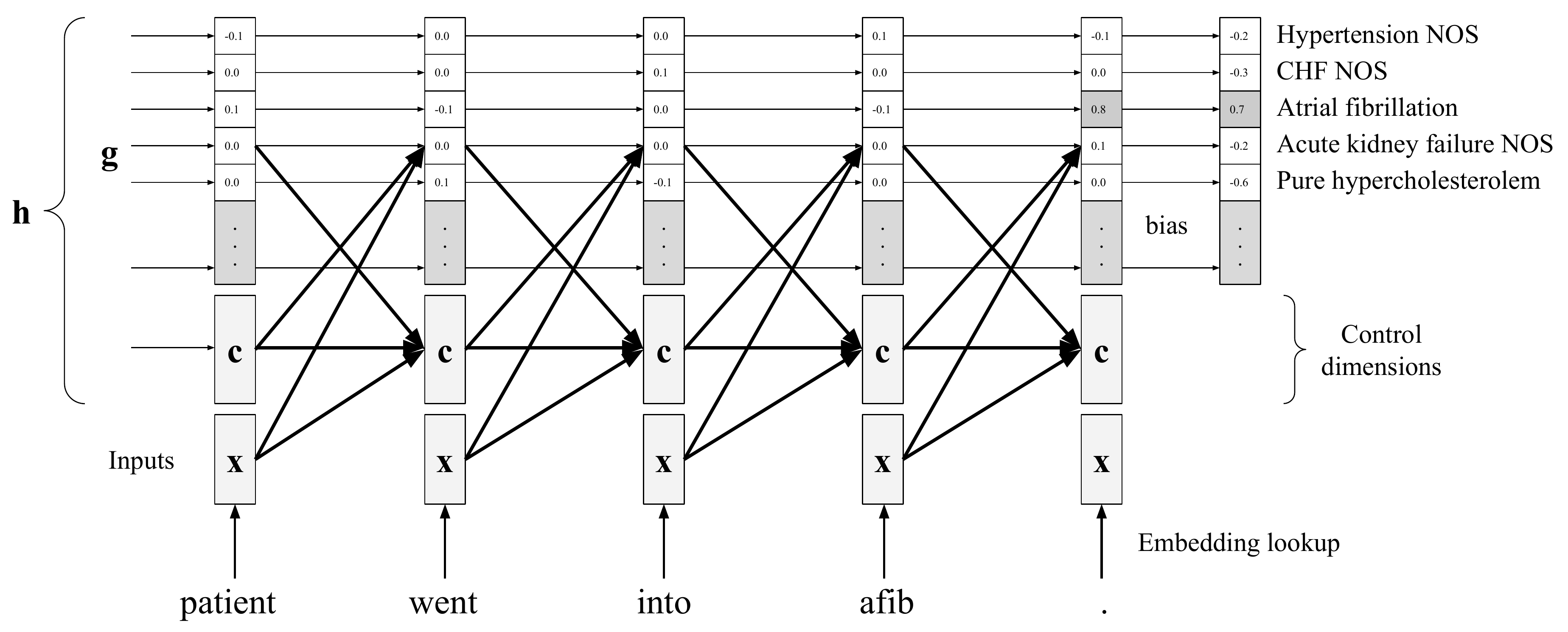}
    \caption{The Grounded RNN architecture. At each time step, the model's belief in the presence of a concept of interest (stored in the corresponding dimension of $\vg$) is updated based on the control part of the recurrent state $\vc$ and the inputs $\vx$.}
    \label{fig:grnn}
\end{figure}

In this section, we show how to derive a Grounded Recurrent Neural Network (GRNN) architecture given a label set $\mathcal{L}$. Note that while we decide to build our version of GRNN on top of a Gated Recurrent Unit, similar ideas can be applied to add grounding to other types of recurrence functions, such as the Elman or LSTM unit.

\paragraph{Sequence Labeling with GRUs} Let $\mathcal{L}$ be a set of labels of interest. Consider a text sequence $\vx = (\vx_1, \ldots, \vx_T)$ and a label assignment for the sequence denoted by $\vy \in \{0, 1\}^{|\mathcal{L}|}$. For ease of exposition, we identify the words $(\vx_1, \ldots, \vx_T)$ with their embeddings of dimension $D_e$. The task of sequence labeling consists in predicting $\vy$ given $\vx$. To that end, we can use a Recurrent Neural Network to obtain a vector representation of the text of dimension $D_h$, then compute the likelihood of each label being present given that representation. More specifically, the recurrent model starts with a representation $\vh_0$, and updates it at each timestep by using a recurrence function $f$ to obtain the global sequence representation $\vh_T$:

\begin{equation}
\forall t \in \{1,\ldots,T\}, \quad \vh_t = f(\vh_{t-1}, \vx_t)
\end{equation}

In the case of the Gated Recurrent Unit, which we build upon in this work, the recurrence function $f$ is parameterized by the $D_h \times (D_e + D_h)$ matrices $\mZ$, $\mR$ and $\mW$ (and dimension $D_h$ bias vectors $\vb_z$, $\vb_r$ and $\vb_w$), and computed as follows.
Let $[\va, \vb]$ denote the concatenation of vectors $\va$ and $\vb$, and $\odot$ be the element-wise product. Then, $\vh_t = f(\vh_{t-1}, \vx_t)$ is given by:

\begin{align}
\vz_t         &= \sigma \left( \mZ [\vh_{t-1},\vx_t] + \vb_z \right) \label{equ:z} \\
\vr_t         &= \sigma \left( \mR [\vh_{t-1},\vx_t] + \vb_r \right) \label{equ:r} \\
\tilde{\vh}_t &= \tanh \left( \mW [\vr_t \odot \vh_{t-1}, \vx_t] + \vb_w \right) \label{equ:ht} \\
\vh_t         &= (1 -\vz_t)\odot \vh_{t-1} + \vz_t \odot \tilde{\vh}_t
\end{align}

From the final text representation $\vh_T$, we can obtain a prediction vector $\tilde{\vy} \in [0, 1]^{|\mathcal{L}|}$ by applying an affine transformation followed by a sigmoid function, parameterized by a $|\mathcal{L}| \times D_h$ matrix $\mP$ and bias vector $\vb_p$. The model parameters $\Theta$ can then be learned by minimizing the expected sum of the binary cross-entropy loss for each (text, label set) pair, denoted as $L$:
\begin{equation}
\forall (\vx, \vy), \quad \tilde{\vy} = \sigma( \mP \vh_T + \vb_p ) \qquad \text{and} \qquad L(\vx,\vy; \Theta) = - \sum_{l=1}^{|\mathcal{L}|} y_l \log (\tilde{y}_l) \label{equ:rnn_proj}
\end{equation}

\paragraph{Grounded Dimensions}
At a high level, the above approach corresponds to having a model summarize all of the relevant information from a text sequence in the final recurrent state $\vh_T$, such that each label corresponds to a different sub-space. This presents several challenges. On the one hand, if the dimension of the recurrent space is much smaller than the label space size, the model capacity might be too small to store the required information about all of the labels in the target set.
On the other hand, too large a recurrent space, while making the model more expressive, can make it prone to over-fitting, rendering optimization difficult when training data is limited. In addition, even though GRUs and LSTMs are better than standard Elman units at modeling long term dependencies, it can still be challenging for them to maintain relevant information from the very beginning of longer sequences (up to a few thousand words in many applications). Our goal in this work is to alleviate the aforementioned problems by adding grounded dimensions to the model's recurrent space.

We split the recurrent state $\vh$ into $|\mathcal{L}|$ \emph{grounded} dimensions $\vg$ and $D_c$ \emph{control} dimensions $\vc$. At each time step, the value stored in a grounded dimension $g_l$ corresponds to the model's current belief that $y_l = 1$. Since by construction we have $\vg \in [-1, 1]^{|\mathcal{L}|}$, the label predictions $\tilde{\vy}$ can simply be obtained by scaling and shifting the final grounded state $\vg_T$. However, we also found it useful to use a bias term to allow grounded dimensions to be centered on 0 regardless of the corresponding label frequency in the data. With this rescaling, and keeping with the notations introduced in the previous paragraph, the model predictions are then given by:
\begin{equation}
\tilde{\vy} = \sigma \big ( \sigma^{-1} \left( \frac{\vg_T + 1}{2} \right) + \vb_p \big )
\end{equation} % also eq to: \sigma ( 2\tanh^{-1} (\vg_T) + \vb_p )
 Figure \ref{fig:grnn} illustrates this process as the model reads the end of a medical note: after reading the phrase ``patient went into afib'', the model increases its belief that this person was diagnosed with Atrial Fibrillation. This formulation already presents some advantages. For example, dedicated dimensions can make learning of long-term dynamics easier. However, simply applying a GRU update to the complete $\vh=[\vg, \vc]$ recurrent state at each time step yields a model with very large capacity ($D_h = |\mathcal{L}| + D_c$), which, as stated previously, can make optimization difficult with limited data. We address this issue in the next paragraph.

\begin{figure}[h]
    \centering
    \includegraphics[width=0.5\textwidth]{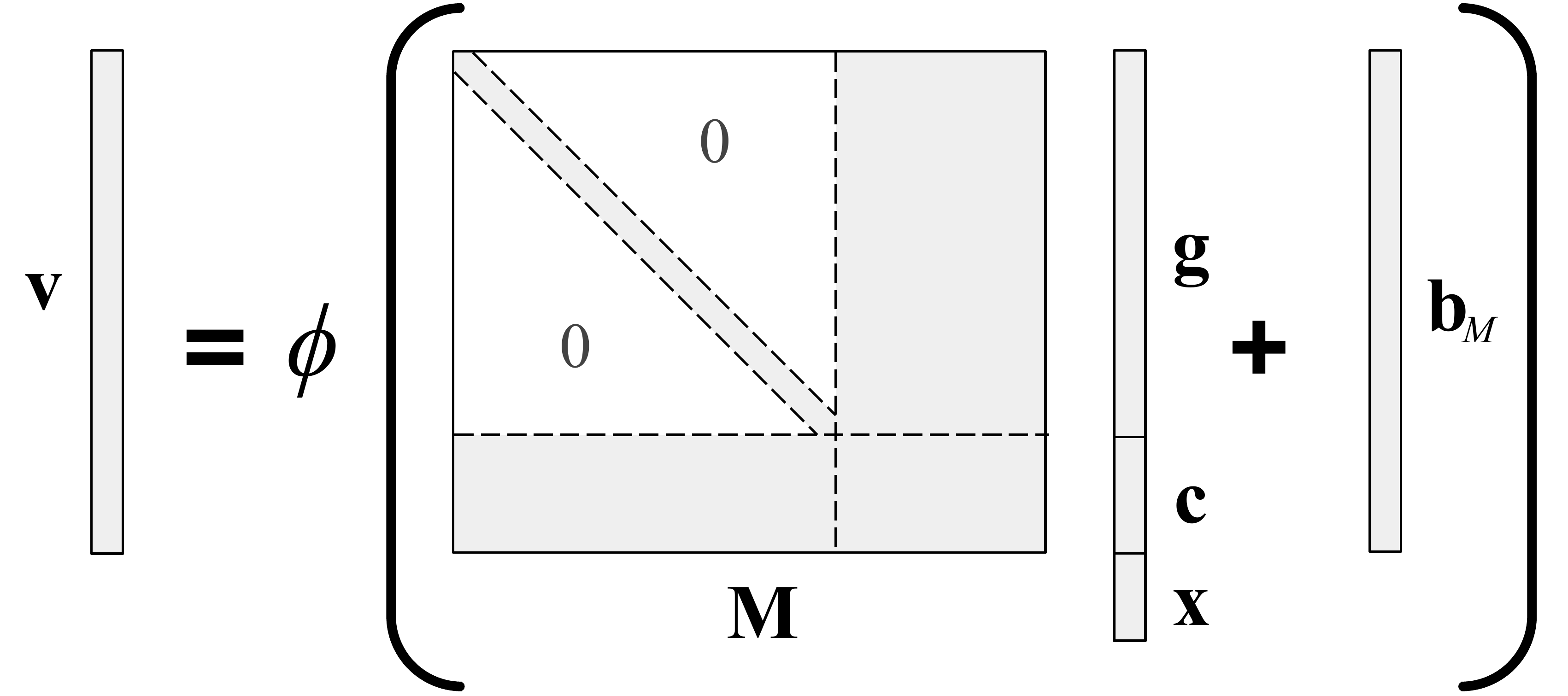} 
    \caption{The semi diagonal transition operation used to update $\vh = [\vg, \vc]$. $\vv$ corresponds to $\vz$, $\vr$ or $\tilde{\vh}$ from Equations \ref{equ:z}, \ref{equ:r} and \ref{equ:ht} respectively. $\mM \in \{\mZ, \mR, \mW\}$ is a weight matrix and $\vb_M$ is the corresponding bias vector.}
    \label{fig:grnn_diag}
\end{figure}

\paragraph{Semi Diagonal Updates}
% \ankit{are we calling this semidiagonal or partially diagonal?}
When given liberty to use the $|\mathcal{L}|$ grounded dimensions without constraints, the model can choose to use them to store other information than label-specific beliefs, especially in the case of dimensions corresponding to rarer concepts. To avoid this potential issue, we propose to restrict the model dynamics by making the $\mZ$, $\mR$ and $\mW$ matrices in Equations \ref{equ:z} to \ref{equ:ht} semi-diagonal, as illustrated in Figure \ref{fig:grnn_diag}. 

With this architecture, the value of any $g_l$ is updated at each timestep based solely on its previous state, the current input and control state $\vc$, which is then made responsible for modeling correlations. Moreover, the number of parameters of the model and computational cost of each time step with semi diagonal transitions grows linearly in the size of the label space, which allows learning to remain tractable even when $|\mathcal{L}|$ is large.

\paragraph{Bidirectional GRNN}
Finally, in many tasks, it is common for bidirectional recurrent models to outperform the unidirectional recurrent models, as they allow for context from the future to be taken into consideration at each timestep along with the history. We extend the GRNN to the bidirectional setting by running a standard GRU in the reverse direction on the document, and concatenating the outputs of this GRU to the inputs of the GRNN, allowing modifications of the grounded dimensions to be based on future context as well as past.

In this Section, we presented the Grounded Recurrent Neural Network: a recurrent architecture designed to have significantly higher capacity than comparable models while making optimization easier and improving interpretability. Section \ref{sec:experiments} analyzes the model's behavior on real world data, and demonstrates these properties experimentally.

\begin{comment}
\subsection{Sliced weights}

In many text classification settings, the number of positive labels far outnumber the number of negative labels for a document. Specifically in multiclass settings, there is exactly one positive label. To speed up GRNN training, we can train the model on subsets of labels and slice the recurrent weight matrices and biases appropriately. For instance, if label $i$ is not to be considered during training, the $i$th row and the $i$th column of the recurrent weight matrix can be sliced out. For a label $i$ that has been sliced out, $g_i$ will be considered to be $0$, and its final label probability would end up being $\sigma(b_i)$, where $b_i$ is the bias applied to the unbiased probability of label $i$.

The selection of labels for slicing can be done based on a fast high-recall first step. We used a bag-of-words logistic regression model to output per-label probabilities for all $k$ labels. The $m$ labels with the highest probabilities from this model are used for the GRNN training. Thus, we can use the GRNN as a fine-tuning step to improve precision of a weaker but faster model.

%Although the bidirectional GRNN's grounded dimensions are less interpretable due to having additional future context, we found that it always performs better than the unidirectional GRNN.
\end{comment}

\section{Experiments}
\label{sec:experiments}

\subsection{Experimental Setting}

\paragraph{Datasets}
We evaluate our model on the task of multi-label classification on three datasets: two versions of the MIMIC medical dataset and Stack Overflow questions.

MIMIC-II \citep{mimic2} is a dataset of Intensive Care Unit medical records. Each patient admission ends with a free text discharge summary describing the patient's stay, diagnoses, and procedures performed. Here, we consider the problem of predicting diagnosis codes from the text, learning from tags manually provided by humans after going through the admission records. We follow \cite{baseline} in extending the label set to also consider parents of the gold label codes in the ICD9 hierarchy, which yields a total vocabulary of $7042$ ICD9 codes that we use as labels, with $36.7$ labels per note on average. We split the data into the same training and test sets as \cite{baseline}, and further split their training set into training and validation. This gives us $18,822$ training, $1711$ validation, and $2282$ testing notes. The average sequence length was $1409.6$, and the maximum length was truncated at $4000$. MIMIC-III \citep{mimic3} is an updated version of the MIMIC-II dataset with thousands more patients and admissions. For this dataset, we predict ICD9 diagnosis as well as procedure codes. We consider the most frequent 4000 diagnoses and 1000 procedures as our label space, giving us about $11.5$ diagnoses and $4.4$ procedures per note on average. We used a training set of $36,998$ notes with $1356$ and $2755$ notes for validation and testing, respectively. The average length for MIMIC-III discharge summaries is $1720.3$.

Stack Overflow\footnote{\url{https://stackoverflow.com/}} is a website which features a large number of computer programming questions and answers. Every question on Stack Overflow has tags defined by the asker. We used a subset of the Stack Overflow data, downloaded from the Kaggle website\footnote{\url{https://www.kaggle.com/stackoverflow/stacksample}} to evaluate the GRNN on the task of predicting tags from question text. We pre-process the data to remove all code blocks from the questions, and select questions which have more than 100 words left. This gave us $365,192$ training, $13,390$ validation and $27,187$ testing samples. We chose the $4000$ most frequent tags as our label set, with $2.9$ tags per sample on average. The average sentence length is much shorter at $190.5$ words, and we truncate the maximum length to $600$.

\paragraph{Baselines}
We first compare the GRNN with a Bag-of-Words baseline, where independent binary classifiers are trained via $L1$-regularized logistic regression for each label, with early stopping based on validation loss. The regularization parameters are tuned independently for each label using the validation sets. Considering recent advances in attention-based models \citep{DBLP:journals/corr/BahdanauCB14}, we also devise a neural Bag-of-Words approach that uses soft attention over the words in a note and adds their embeddings. For all our neural approaches, we used a word embedding size of $192$. The attention scores for words are based on the local neighborhoods of the words, and are shared by all labels. This method ensures that only the significant words are considered for the prediction of labels. The note representation $\vh$ is then computed as the weighted sum of the embeddings, and we obtain predictions and learn the model parameters as described in Section \ref{sec:grnn} (replacing $\vh_T$ with $\vh$ in Equation \ref{equ:rnn_proj}). The GRU baseline is as defined in Section \ref{sec:grnn}, with a hidden size of either $128$, or a larger dimension corresponding to a GRU with the same number of parameters as the corresponding GRNN ($846$ for $5000$ labels, $793$ for $4000$). This allows us to test whether the difference in our model's performance is due to grounding or simply comes from an increased capacity. We also consider bidirectional GRUs with a dimension $64$ hidden state. Furthermore, for the MIMIC-II dataset, we compare our results to those of \cite{baseline}, which uses a flat and a hierarchical SVM for the same task. For each example, the hierarchical SVM node decision for an ICD9 code is trained only if its parent code is positive, and a child code is evaluated only if its parent is classified as positive during testing. The flat SVM predicts all the leaf ICD9 codes independently and builds the extended predictions according to the hierarchy. Unlike \cite{baseline}, we learn and predict on the entire extended label set without considering the ICD9 hierarchy, letting our learning algorithm infer relevant label correlations. However, we also note that the authors of the baseline SVM models could have obtained better results by fine-tuning their regularization parameters.

\paragraph{Other Comparisons}
We ran early experiments with grounded models without a semi diagonal constraint on the recurrent transition matrices, and found that such models failed to generalize for large label spaces due to over-fitting. We also tried grounded recurrent models without any control dimensions, which always performed worse, since we deny the model the capacity to track history beyond current beliefs in labels. In addition, we investigated a variant of the bidirectional GRU closer to our formulation of the bidirectional GRNN, wherein the outputs of a reverse GRU are concatenated to the inputs for a forward GRU (denoted as BiGRU-l in the supplementary material tables), but we found that it always performed worse than the standard bidirectional GRU described earlier. 

Finally, we ran the entity network of \citep{henaff2016tracking} on MIMIC-III text with $6$ entity RNNs. The keys for the entities are learned globally, enabling the network to learn clusters of related labels, with each entity tracking one such cluster. For predictions, each label performed attention over the entity network blocks to determine its value, enabling each label to focus on the cluster it is a part of. This network took several days to train while performing similar to or worse than the GRU baselines.

\paragraph{Curriculum Learning}
To maintain the invariance of the grounded dimensions representing the likelihoods of labels at intermediate timesteps, we train the model on truncated documents. A convenient way to do this is to start with small sentence lengths and increase the maximum document length as training progresses. At smaller lengths, this helps learn the overall statistics of labels and any early evidence in text. As the maximum document length is increased, the model learns to attribute labels with evidence in text that appears later on. Since we never go back to a smaller length, this enables the model to fine-tune its predictions as more useful information becomes available. We can also view this strategy as a way of doing curriculum learning, enabling the model to perform well on long documents by initially learning on shorter documents. The initial document length was set to $50$, and increased by a factor of $1.35$ on every training epoch. Initial experiments without curriculum learning took much longer to train, and performed worse than models trained with curriculum learning. We found that this approach got better results and trained faster for all recurrent models, and thus we decided to use the same strategy for the GRU baselines as well.

\subsection{Concept Extraction Results}

\begin{table}[t!]
\tiny
\begin{centering}
\begin{tabular}{l c c c c c c c c c c c c c c}
\toprule
Model     & P     & R     & &  \multicolumn{2}{c}{F1}& \multicolumn{2}{c}{AUC(PR)} & \multicolumn{2}{c}{AUC(ROC)} & & \multicolumn{2}{c}{P@$n$} & \multicolumn{2}{c}{R@$n$} \\
          &       &       & & Micro & Macro & Micro & Macro & Micro & Macro & &  8    &  40   &   8   &  40   \\
\midrule
Flat SVM$^{*}$  & \bf 0.867 & 0.164 & & 0.276 & - &   -   &   -   &   -   &   -   & &   -   &   -   &   -   &   -   \\
Hier. SVM$^{*}$ & 0.577 & 0.301 & & 0.395 &   -   &   -   &   -   &   -   &   -   & &   -   &   -   &   -   &   -   \\
\midrule
Logistic  & 0.774 & 0.395 & & 0.523 & 0.042 & 0.541 & 0.125 & 0.919 & 0.704 & & 0.913 & 0.572 & 0.169 & 0.528 \\
Attn BoW  & 0.745 & 0.399 & & 0.520 & 0.027 & 0.521 & 0.079 & 0.975 & 0.807 & & 0.912 & 0.549 & 0.169 & 0.508 \\
GRU-128   & 0.725 & 0.396 & & 0.512 & 0.027 & 0.523 & 0.082 & \bf 0.976 & \bf 0.827 & & 0.906 & 0.541 & 0.168 & 0.501 \\
BiGRU-64  & 0.715 & 0.367 & & 0.485 & 0.021 & 0.493 & 0.071 & 0.973 & 0.811 & & 0.892 & 0.522 & 0.165 & 0.483 \\
\midrule
GRNN-128  & 0.753 & \bf 0.472 & & \bf 0.580 & 0.052 & 0.587 & 0.126 & \bf 0.976 & 0.815 & & \bf 0.930 & 0.592 & \bf 0.172 & 0.548 \\  
BiGRNN-64 & 0.761 & 0.466 & & 0.578 & \bf 0.054 & \bf 0.589 & \bf 0.131 & 0.975 & 0.798 & & 0.925 & \bf 0.596 & \bf 0.172  & \bf 0.552 \\
\bottomrule
\end{tabular}
\vspace{0.1in}
\caption{Results on MIMIC-2, 7042 labels. $^{*}$ Lines are taken from \citep{baseline}.}
\label{tab:mimic2}
\end{centering}
\end{table}

\begin{table}[t!]
\small
\begin{centering}
\begin{tabular}{l c c c c c c c c c c c c c c}
\toprule
Model     &  \multicolumn{2}{c}{F1}& \multicolumn{2}{c}{AUC(PR)} & \multicolumn{2}{c}{AUC(ROC)} & & \multicolumn{2}{c}{P@$n$} & \multicolumn{2}{c}{R@$n$} \\
          & Micro & Macro & Micro & Macro & Micro & Macro & &  8    &  40   &   8   &  40   \\
\midrule
Logistic  & 0.432 & 0.063 & \bf 0.441 & \bf 0.156 & 0.930 & 0.780 & & 0.612 & 0.252 & 0.309 & 0.638 \\
Attn BoW  & 0.406 & 0.057 & 0.362 & 0.096 & 0.969 & 0.873 & & 0.578 & 0.237 & 0.292 & 0.600 \\
GRU-64    & 0.351 & 0.045 & 0.347 & 0.100 & 0.970 & \bf 0.885 & & 0.524 & 0.223 & 0.265 & 0.565 \\
GRU-128   & 0.400 & 0.057 & 0.376 & 0.104 & 0.970 & 0.876 & & 0.570 & 0.232 & 0.289 & 0.587 \\
GRU-846   & 0.384 & 0.059 & 0.325 & 0.090 & 0.960 & 0.839 & & 0.522 & 0.215 & 0.264 & 0.543 \\
BiGRU-64  & 0.382 & 0.052 & 0.360 & 0.101 & 0.969 & 0.871 & & 0.551 & 0.227 & 0.279 & 0.575 \\
\midrule
GRNN-128  & 0.456 & 0.072 & 0.412 & 0.119 & 0.967 & 0.857 & & 0.622 & 0.249 & 0.315 & 0.631 \\  
BiGRNN-64 & \bf 0.464 & \bf 0.074 & \bf 0.441 & 0.131 & \bf 0.972 & 0.872 & & \bf 0.633 & \bf 0.258 & \bf 0.321 & \bf 0.654 \\
\bottomrule
\end{tabular}
\vspace{0.1in}
\caption{Results on MIMIC-3, 5000 labels. }
\vspace{-0.25in}
\label{tab:mimic3}
\end{centering}
\end{table}

\paragraph{Quantitative Evaluation}
Tables \ref{tab:mimic2}, \ref{tab:mimic3} and \ref{tab:stack} report quantitative results of our model and baselines on MIMIC-II, MIMIC-III, and StackOverflow data respectively. We report the F1 measure, and Area Under the Precision/Recall (AUC(PR)) and ROC (AUC(ROC)) curves. The Micro-averaged versions of the measures correspond to considering any (text, label) pair as an independent prediction, either true or false, and computing the statistics on all of those together. To obtain macro-averaged measures, the statistics are computed independently on each label, then averaged uniformly, regardless of the label frequency in the data. Compared to the micro-averaged versions, macro-averaging puts much more of a weight on the model's ability to accurately predict rare labels. We also consider our model's performance in actual health care applications. Given the specific requirements of the domain, one successful human-in-the-loop strategy consists in using the model scores to show a user the $n$ highest scored labels in a highly multi-class application, or to use these scores to improve an auto-complete system as in \cite{jernite2013predicting} and \cite{greenbaum2017contextual}. In that case, it is important to know how many of the proposals are correct (precision at $n$: P@$n$ in the tables). Indeed, low precision can significantly hurt a user's confidence in the system. Additionally, we want to know how many of the example's labels are covered by the proposed predictions (recall at $n$: R@$n$). We provide these measures for $n=8$ and $n=40$.

Tables \ref{tab:mimic2} and \ref{tab:mimic3} show that the GRNN performs significantly better on medical data all measures combined. The gap is greater for MIMIC-II, which agrees with our intuition: MIMIC-II has less data, which makes having a data efficient model more important, and the target label space has a hierarchical structure, which makes being able to take advantage of correlations all the more useful. In particular, the advantage of the grounded architecture is most noticeable on the precision and recall at $n$ measures, which correspond to our proposed use case. Finally, we give results on StackOverflow data in Table \ref{tab:stack} to show that our model also performs well on a different domain and setting: the dataset is much larger, while the individual example text sequences are significantly shorter. The grounded architectures are on par with the best baselines, and still consistently out-perform them on the P@$n$ and R@$n$ measures.

\begin{figure}[t!]
    \centering
    \includegraphics[width=0.393\textwidth]{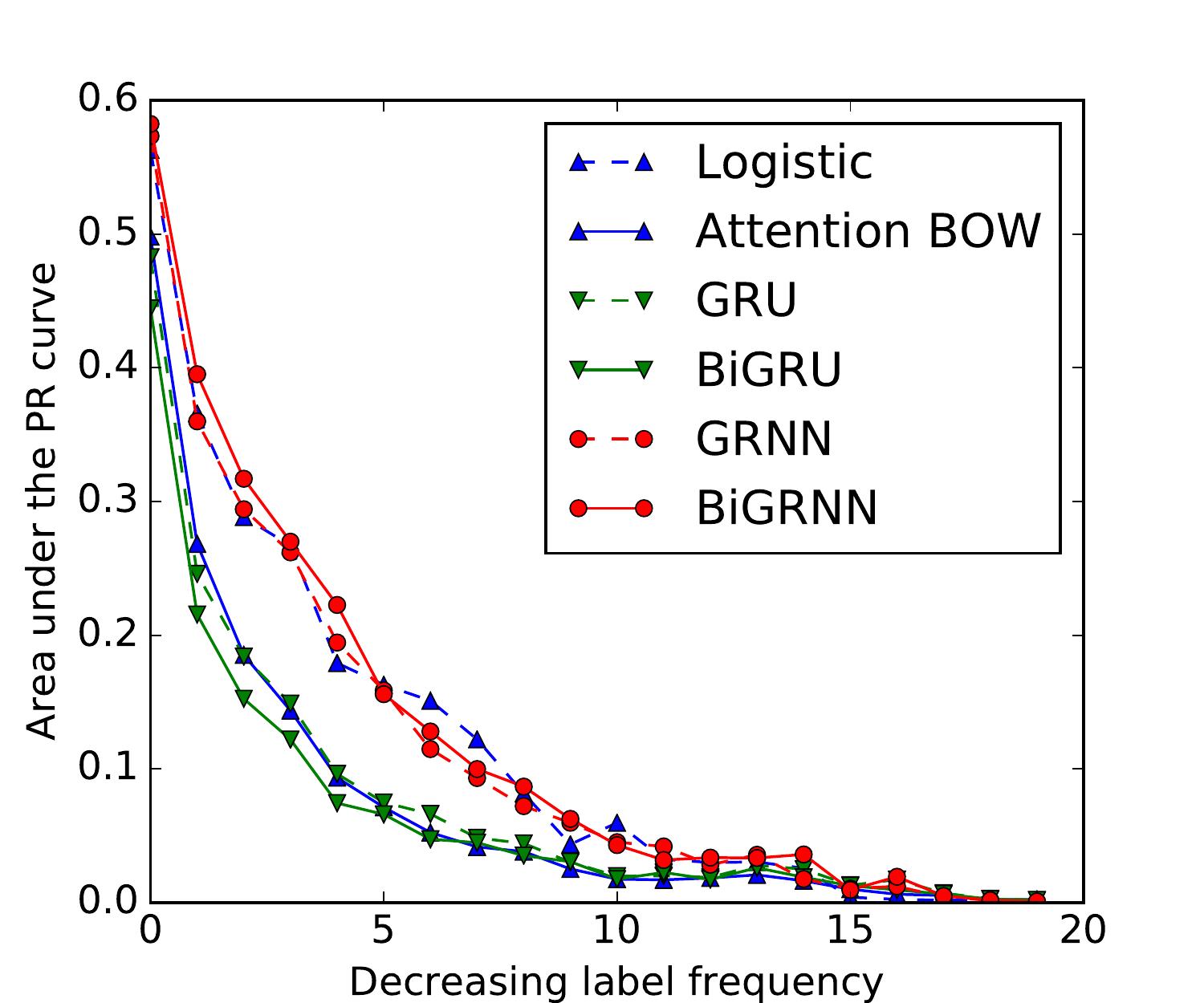}
    \includegraphics[width=0.48\textwidth]{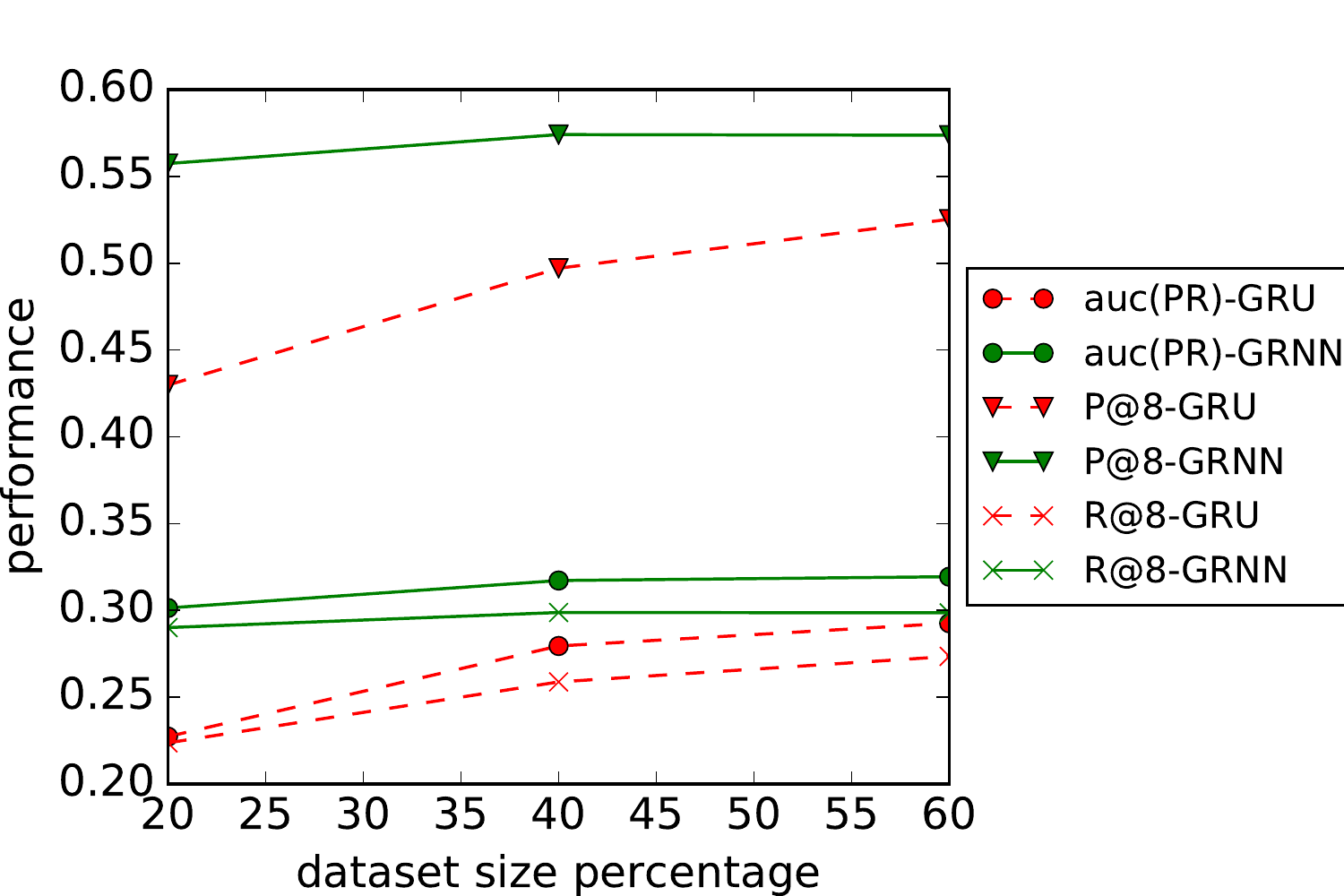}
    \caption{{\bf Left:} Per-label AUC(PR) as a function of the label frequency on MIMIC-II. The labels are ordered from most to least frequent. {\bf Right:} Comparing the performance of the GRU and GRNN by varying the amount of training data available.}
    \label{fig:roc_labelfreq}
\end{figure}

\paragraph{Model Introspection}
Figure \ref{fig:roc_labelfreq} provides more insight into the properties of the model architecture. The left plot shows that the GRNN AUC(PR) outperforms most baselines regardless of label frequency. We note that the Logistic curve is close to the grounded architectures, which corresponds to the similar macro-averaged PR AUC, as shown in Table \ref{tab:mimic2}. Compared to the other models, the GRNN has most of an advantage on concepts in the middle of the frequency spectrum: labels which appear between a few dozen and a few hundred times in the training data. We also investigate our model's data efficiency by training both a GRU and GRNN on subsets of MIMIC-III of increasing size: using 20\%, 40\% and 60\% of the data. While the GRNN always outperforms the standard GRU, the gap is larger the less training data the model has access to, which implies that grounding does indeed allow a recurrent neural network to learn from less data.

\begin{table}[t!]
\small
\begin{centering}
\begin{tabular}{l c c c c c c c c c  c c c c}
\toprule
Model  & \multicolumn{2}{c}{F1}  & \multicolumn{2}{c}{AUC(PR)} & \multicolumn{2}{c}{AUC(ROC)} & & \multicolumn{2}{c}{P@$n$} & \multicolumn{2}{c}{R@$n$} \\
             & Micro & Macro & Micro & Macro & Micro & Macro & &   8   &  40   &   8   &  40  \\
\midrule
Logistic     & 0.495 & 0.157 & 0.497 & 0.206 & 0.951 & 0.808 & & 0.253 & 0.059 & 0.707 & 0.827 \\
Attn BoW     & 0.549 & 0.169 & 0.517 & 0.196 & \bf 0.990 & 0.975 & & \bf 0.271 & \bf 0.064 & \bf 0.760 & 0.897 \\
GRU-128      & 0.552 & 0.191 & 0.532 & \bf 0.231 & 0.989 & 0.972 & & 0.263 & 0.063 & 0.738 & 0.880 \\
GRU-793      & 0.545 & \bf 0.204 & 0.505 & 0.215 & 0.989 & 0.971 & & 0.263 & 0.063 & 0.738 & 0.882 \\
BiGRU-64     & 0.554 & 0.189 & 0.531 & 0.221 & 0.991 & \bf 0.976 & & 0.269 & \bf 0.064 & 0.753 & 0.893 \\
\midrule
GRNN-128     & 0.539 & 0.153 & 0.514 & 0.190 & 0.985 & 0.957 & & 0.265 & 0.063 & 0.742 & 0.884 \\  
BiGRNN-64    & \bf 0.560 & 0.179 & \bf 0.536 & 0.214 & 0.989 & 0.973 & & \bf 0.271 & \bf 0.064 & \bf 0.760 & \bf 0.902 \\
\bottomrule
\end{tabular}
\vspace{0.1in}
\caption{Results on StackOverflow, 4000 labels.}
\label{tab:stack}
\end{centering}
\end{table}

\paragraph{Interpretable Predictions}
We also want our model to provide interpretable predictions. Indeed, better interpretability of the model decisions is as important as improved quantitative performances in a medical setting, where practitioners need to be able to trust the system they use, and to easily query the decision process for predictions that are more surprising to them. It should be noted that one can obtain some limited insight into the decision process of the attention-based baseline, for instance, by looking at the global scores. However, for both the GRU and GRNN, we can actually track the model's belief in the presence of a \emph{specific label} as a note is read. As mentioned in Section \ref{sec:grnn}, for the GRNN, one simply needs to look at the evolution of the corresponding grounded dimension between $-1$ and $1$. It is possible to obtain similar information from the GRU, by applying the projection defined in Equation \ref{equ:rnn_proj} at each time step, which gives a value between $0$ and $1$.

\begin{figure}[t!]
    \centering
    \includegraphics[width=0.495\textwidth]{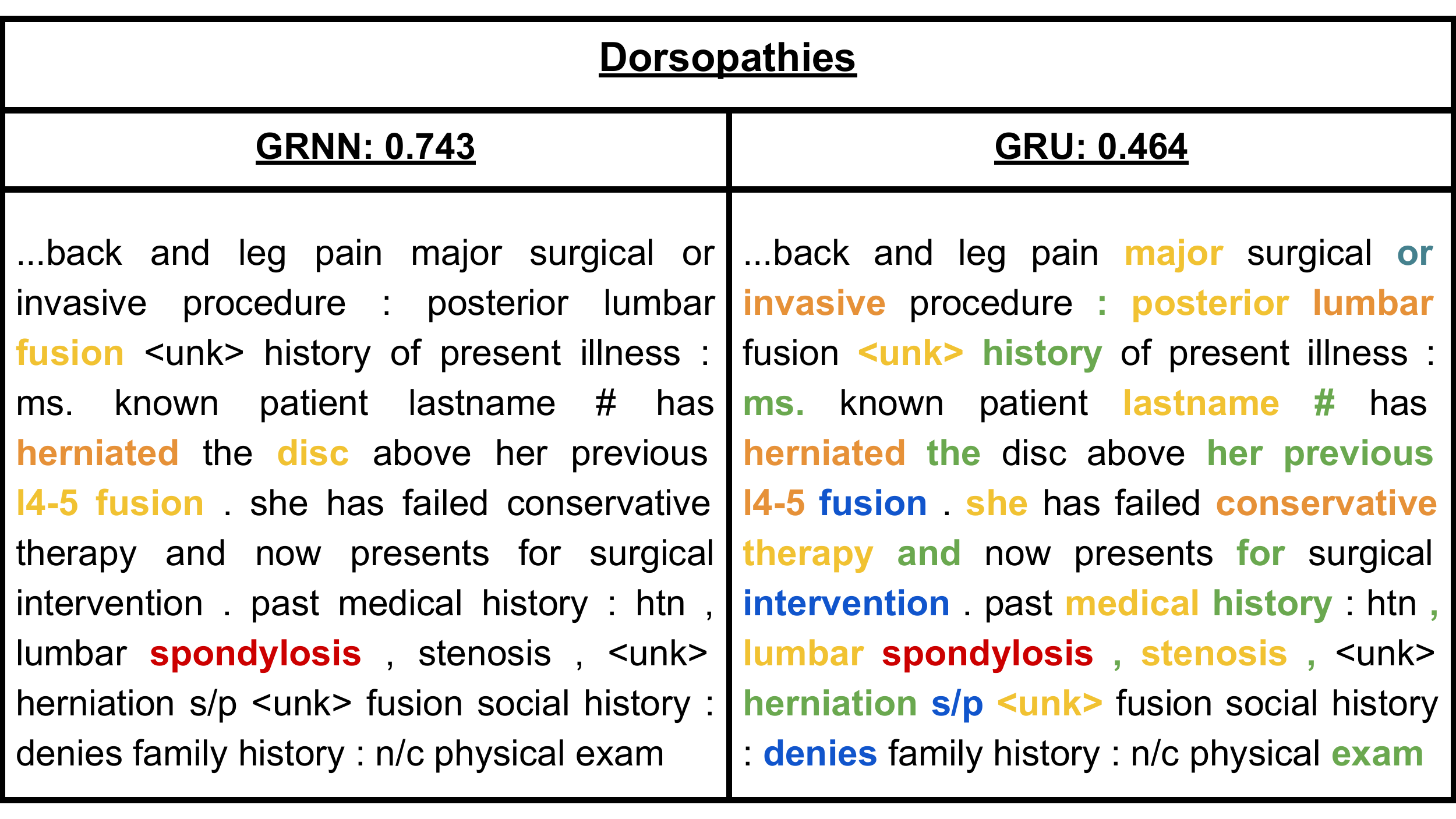}
    \includegraphics[width=0.495\textwidth]{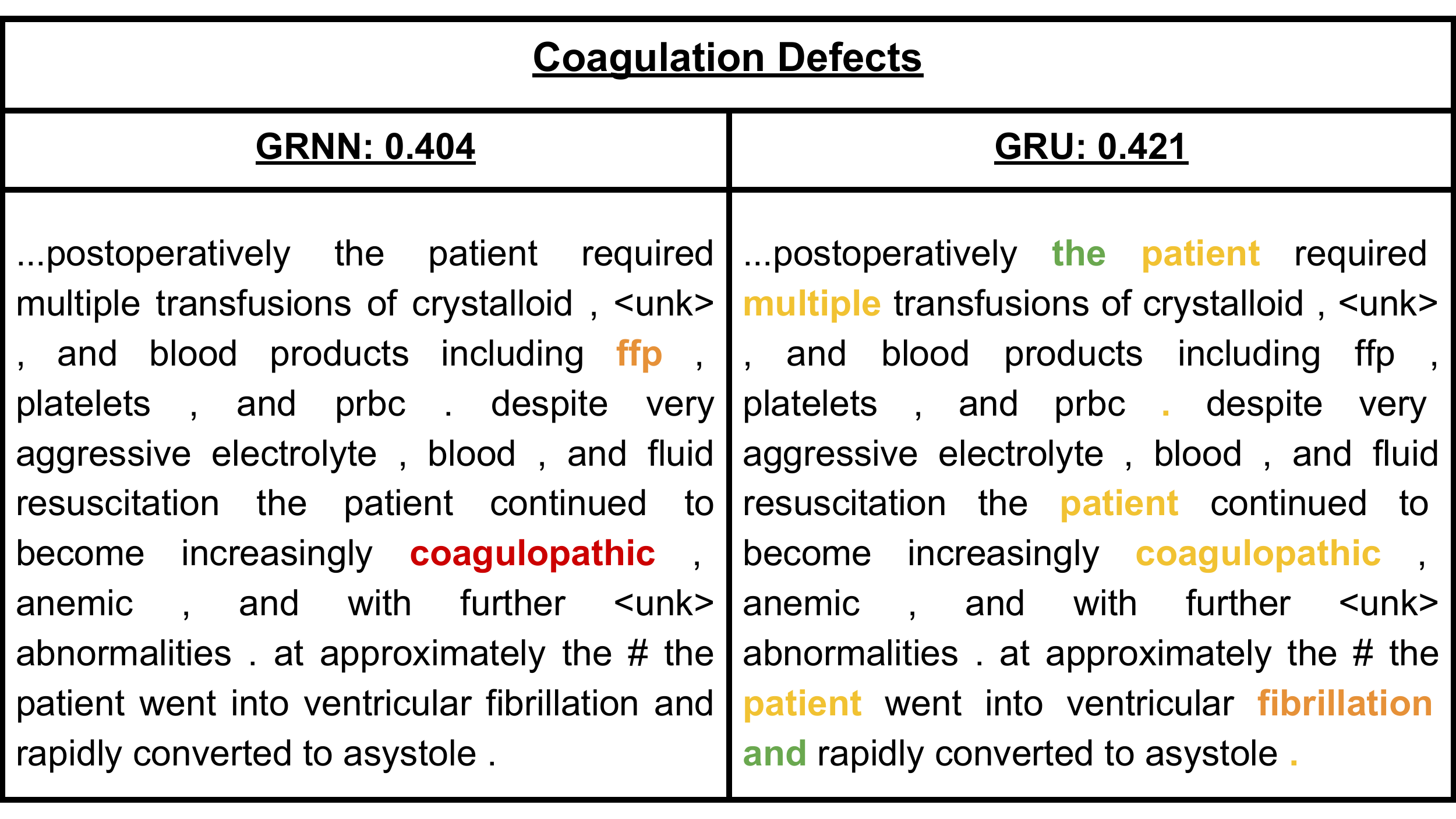}
    \caption{Evolution of the network belief state while reading a note. The color scale from blue to red indicates whether the belief decreases or increases respectively at each time step.}
    \label{fig:activations}
\end{figure}

Figure \ref{fig:activations} presents this visualization for extracts from two discharge summaries, outlining the time steps which either increase (red, orange, yellow) or decrease (green or blue) the model's belief in the presence of a concept. In both cases, the Grounded architecture provides a sharper, more interpretable signal, focusing on clinically meaningful passages (``herniated disc'' or ``lumbar spondylosis'' for the patient with dorsopathies, ``transfusions of [\ldots] ffp'' and ``become increasingly coagulopathic'' for the coagulation defect diagnosis). On the other hand, while the GRU's belief does increase somewhat on those same phrases, this effect is similar to that of other more distantly related phrases (``fibrillation'') as well as some that do not seem especially relevant (``patient'').

\section{Conclusion}
\label{sec:conclusion}
In this work, we introduce the Grounded Recurrent Neural Network, a recurrent network architecture which learns to perform multi-label text classification in a data efficient way by tying concepts of interest to specific dimensions of its hidden state. At the same time structural constraints on the recurrence matrices allow the model to remain tractable even in the presence of a large number of labels. Thus, the model is able to combine the data efficiency of simple Bag-of-Word text classification methods with an RNN's ability to model linguistic structures by tying labels of interest to specific dimensions of its hidden state.

We show that our model is especially suited to a medical setting where its ability to learn from limited data, model concept correlations, and provide interpretable predictions lead to both better performance and improved trustworthiness for practitioners over several strong baselines. We also demonstrate our network's ability to match or outperform these baselines even in a case where data efficiency is less crucial. We hope to improve our model further in future work by introducing structured prediction objectives so as to take better advantage of our proposed architecture's ability to represent interactions between concepts.

\subsubsection*{Acknowledgments}
Yacine Jernite and David Sontag gratefully acknowledge the support of the Defense Advanced Research Projects Agency (DARPA) Probabilistic Programming for Advancing Machine Learning (PPAML) Program under Air Force Research Laboratory (AFRL) prime contract no. \texttt{FA8750-14-C-0005}. Any opinions, findings, and conclusions or recommendations expressed in this material are those of the author(s) and do not necessarily reflect the view of DARPA, AFRL, or the US government.

\medskip

\small
\bibliography{bibliography}

\clearpage

\appendix
\section*{Supplement to:\\ Grounded Recurrent Neural Networks\\}

\subsection*{Experiments: Full Tables}

\begin{table}[h]
\tiny
\begin{centering}
\begin{tabular}{l c c c c c c c c c c c c c c c c}
\toprule
Model     & P     & R     & &  \multicolumn{2}{c}{F}& \multicolumn{2}{c}{AUC(PR)} & \multicolumn{2}{c}{AUC(ROC)} & & \multicolumn{3}{c}{P@$n$} & \multicolumn{3}{c}{R@$n$} \\
          &       &       & & Micro & Macro & Micro & Macro & Micro & Macro & &  8    &  24   &  40   &   8   &  24   &  40   \\
\midrule
Flat SVM  & 0.867 & 0.164 & & 0.276 & -     &   -   &   -   &   -   &   -   & &   -   &   -   &   -   &   -   &   -   &   -   \\
Hier. SVM & 0.577 & 0.301 & & 0.395 &   -   &   -   &   -   &   -   &   -   & &   -   &   -   &   -   &   -   &   -   &   -   \\
\midrule
Logistic  & 0.774 & 0.395 & & 0.523 & 0.042 & 0.541 & 0.125 & 0.919 & 0.704 & & 0.913 & 0.715 & 0.572 & 0.169 & 0.396 & 0.528 \\
Attn BoW  & 0.745 & 0.399 & & 0.520 & 0.027 & 0.521 & 0.079 & 0.975 & 0.807 & & 0.912 & 0.697 & 0.549 & 0.169 & 0.387 & 0.508 \\
GRU-128   & 0.725 & 0.396 & & 0.512 & 0.027 & 0.523 & 0.082 & 0.976 & 0.827 & & 0.906 & 0.689 & 0.541 & 0.168 & 0.383 & 0.501 \\
BiGRU-64  & 0.715 & 0.367 & & 0.485 & 0.021 & 0.493 & 0.071 & 0.973 & 0.811 & & 0.892 & 0.664 & 0.522 & 0.165 & 0.369 & 0.483 \\
BiGRU-l-64& 0.756 & 0.321 & & 0.443 & 0.013 & 0.457 & 0.053 & 0.971 & 0.800 & & 0.892 & 0.627 & 0.490 & 0.161 & 0.349 & 0.454 \\
\midrule
GRNN-128  & 0.753 & 0.472 & & 0.580 & 0.052 & 0.587 & 0.126 & 0.976 & 0.815 & & 0.930 & 0.744 & 0.592 & 0.172 & 0.414 & 0.548 \\  
GRNN-128b & 0.775 & 0.429 & & 0.552 & 0.042 & 0.579 & 0.123 & 0.977 & 0.802 & & 0.923 & 0.731 & 0.588 & 0.171 & 0.406 & 0.545 \\ 
BiGRNN-64 & 0.761 & 0.466 & & 0.578 & 0.054 & 0.589 & 0.131 & 0.975 & 0.798 & & 0.925 & 0.746 & 0.596 & 0.172 & 0.415 & 0.552 \\
\bottomrule
\end{tabular}
\caption{Results on MIMIC-2, 7042 labels.}
\end{centering}
\end{table}

\begin{table}[h]
\tiny
\begin{centering}
\begin{tabular}{l c c c c c c c c c c c c c c c c}
\toprule
Model  & P & R &  &  \multicolumn{2}{c}{F}& \multicolumn{2}{c}{AUC(PR)} & \multicolumn{2}{c}{AUC(ROC)} & & \multicolumn{3}{c}{P@$n$} & \multicolumn{3}{c}{R@$n$} \\
          &       &       & & Micro & Macro & Micro & Macro & Micro & Macro & &  8    &  24   &  40   &   8   &  24   &  40   \\
\midrule
Logistic  & 0.667 & 0.320 & & 0.432 & 0.063 & 0.441 & 0.156 & 0.930 & 0.780 & & 0.612 & 0.360 & 0.252 & 0.309 & 0.545 & 0.638 \\
Attn BoW  & 0.615 & 0.303 & & 0.406 & 0.057 & 0.362 & 0.096 & 0.969 & 0.873 & & 0.578 & 0.335 & 0.237 & 0.292 & 0.508 & 0.600 \\
GRU-64    & 0.610 & 0.246 & & 0.351 & 0.045 & 0.347 & 0.100 & 0.970 & 0.885 & & 0.524 & 0.310 & 0.223 & 0.265 & 0.470 & 0.565 \\
GRU-128   & 0.629 & 0.293 & & 0.400 & 0.057 & 0.376 & 0.104 & 0.970 & 0.876 & & 0.570 & 0.326 & 0.232 & 0.289 & 0.495 & 0.587 \\
GRU-846   & 0.528 & 0.302 & & 0.384 & 0.059 & 0.325 & 0.090 & 0.960 & 0.839 & & 0.522 & 0.299 & 0.215 & 0.264 & 0.454 & 0.543 \\
BiGRU-64  & 0.619 & 0.277 & & 0.382 & 0.052 & 0.360 & 0.101 & 0.969 & 0.871 & & 0.551 & 0.483 & 0.227 & 0.279 & 0.483 & 0.575 \\
BiGRU-l-64& 0.587 & 0.213 & & 0.312 & 0.031 & 0.302 & 0.075 & 0.966 & 0.869 & & 0.482 & 0.431 & 0.207 & 0.244 & 0.431 & 0.524 \\
\midrule
GRNN-128  & 0.653 & 0.350 & & 0.456 & 0.072 & 0.412 & 0.119 & 0.967 & 0.857 & & 0.622 & 0.356 & 0.249 & 0.315 & 0.541 & 0.631 \\  
GRNN-128b & 0.700 & 0.254 & & 0.373 & 0.033 & 0.379 & 0.095 & 0.975 & 0.899 & & 0.592 & 0.353 & 0.251 & 0.300 & 0.536 & 0.636 \\
BiGRNN-64 & 0.678 & 0.353 & & 0.464 & 0.074 & 0.441 & 0.131 & 0.972 & 0.872 & & 0.633 & 0.368 & 0.258 & 0.321 & 0.561 & 0.654 \\
\bottomrule
\end{tabular}
\caption{Results on MIMIC-3, 5000 labels. }
\end{centering}
\end{table}

\begin{table}[h]
\tiny
\begin{centering}
\begin{tabular}{l c c c c c c c c c c c c c c}
\toprule
Model  & P & R & & \multicolumn{2}{c}{F}  & \multicolumn{2}{c}{AUC(PR)} & \multicolumn{2}{c}{AUC(ROC)} & & \multicolumn{2}{c}{P@$n$} & \multicolumn{2}{c}{R@$n$} \\
             &       &       & & Micro & Macro & Micro & Macro & Micro & Macro & &  8    & 40    &  8    &   40  \\
\midrule
Logistic     & 0.690 & 0.385 & & 0.495 & 0.157 & 0.497 & 0.206 & 0.951 & 0.808 & & 0.253 & 0.059 & 0.707 & 0.827 \\
Attn BoW     & 0.715 & 0.446 & & 0.549 & 0.169 & 0.517 & 0.196 & 0.990 & 0.975 & & 0.271 & 0.064 & 0.760 & 0.897 \\
GRU-128      & 0.689 & 0.460 & & 0.552 & 0.191 & 0.532 & 0.231 & 0.989 & 0.972 & & 0.263 & 0.063 & 0.738 & 0.880 \\
GRU-793      & 0.681 & 0.454 & & 0.545 & 0.204 & 0.505 & 0.215 & 0.989 & 0.971 & & 0.263 & 0.063 & 0.738 & 0.882 \\
BiGRU-64     & 0.695 & 0.461 & & 0.554 & 0.189 & 0.531 & 0.221 & 0.991 & 0.976 & & 0.269 & 0.064 & 0.753 & 0.893 \\
BiGRU-l-64   & 0.692 & 0.412 & & 0.522 & 0.144 & 0.502 & 0.188 & 0.988 & 0.968 & & 0.254 & 0.061 & 0.710 & 0.860 \\
\midrule
GRNN-128     & 0.718 & 0.431 & & 0.539 & 0.153 & 0.514 & 0.190 & 0.985 & 0.957 & & 0.265 & 0.063 & 0.742 & 0.884 \\  
BiGRNN-64    & 0.714 & 0.460 & & 0.560 & 0.179 & 0.536 & 0.214 & 0.989 & 0.973 & & 0.271 & 0.064 & 0.760 & 0.902 \\
\bottomrule
\end{tabular}
\caption{Results on StackOverflow, 4000 labels.}
\end{centering}
\end{table}

\end{document}